\documentclass[
]{ceurart}

\begin{document}

\copyrightyear{2021}
\copyrightclause{Copyright for this paper by its authors.
  Use permitted under Creative Commons License Attribution 4.0
  International (CC BY 4.0).}

\conference{FIRE'21: Forum for Information Retrieval Evaluation, December 13-17, 2021, India}

\title{Exploring Transformer Based Models to Identify Hate Speech and Offensive Content in English and Indo-Aryan Languages}

\author[1]{Somnath Banerjee}[
orcid=0000-0002-9445-8439,
email=som.iitkgpcse@kgpian.iitkgp.ac.in,
]

\author[2]{Maulindu Sarkar}[%
email=moulindu.sarkar@iitkgp.ac.in,
]

\author[2]{Nancy Agrawal}[%
email=nancy08@iitkgp.ac.in,
]

\author[1]{Punyajoy Saha}[%
orcid=0000-0002-3952-2514,
email=punyajoys@iitkgp.ac.in,
]

\author[1]{Mithun Das}[%
orcid=0000-0003-1442-312X,
email=mithundas@iitkgp.ac.in,
]

\address[1]{Department of Computer Science and Engineering, Indian Institute of Technology, Kharagpur, West Bengal, India}

\address[2]{Department of Electrical Engineering , Indian Institute of Technology, Kharagpur, West Bengal, India}

\begin{abstract}
Hate speech is considered to be one of the major issues currently plaguing online social media. Repeated and repetitive exposure to hate speech has been shown to create physiological effects on the target users. Thus, hate speech, in all its forms, should be addressed on these platforms in order to maintain good health. In this paper, we explored several Transformer based machine learning models for the detection of hate speech and offensive content in English and Indo-Aryan languages at FIRE 2021. We explore several models such as mBERT, XLMR-large, XLMR-base by team name "Super Mario". Our models came \(2nd\) position in Code-Mixed Data set (Macro F1: 0.7107), \(2nd\) position in Hindi two-class classification (Macro F1: 0.7797), \(4th\) in English four-class category (Macro F1: 0.8006) and \(12th\) in English two-class category (Macro F1: 0.6447). We have made our code public \footnote{\url{https://github.com/hate-alert/OffensiveLangDetectIndoAryan}}.

\end{abstract}


\maketitle

\section{Introduction}

Online Social media platforms such as Twitter, Facebook have connected billions of people and allowed them to publish their ideas and opinions instantly. The problem arises when the bad actors(users) share contents to spread propaganda, fake news, and hate speech etc\cite{,hate-speech-websci-19} by using these platforms. Companies like Facebook have been accused of instigating anti-Muslim mob violence in Sri Lanka that left three people dead~\footnote{~\url{http://www.aaiusa.org/unprecedented_increase_expected_in_upcoming_fbi_hate_crime_report}} and a UN report blamed them for playing a leading role in the possible genocide of the Rohingya community in Myanmar by spreading hate speech~\footnote{~\url{https://www.reuters.com/investigates/special-report/myanmar-facebook-hate}}.
In order to mitigate the spread of hateful/offensive content, these platforms have come up with some guidelines\footnote{\label{twitter_violation}\url{https://help.twitter.com/en/rules-and-policies/hateful-conduct-policy}} and expect that users should follow the guidelines before sharing any content. Sometimes, violation of such procedures could lead to the post being deleted or user account suspension.

To reduce the harmful content (such as offensive/hate speech) from these platforms, they employ 
moderators~\cite{newton_2019} to keep the conversations healthy and people-friendly by manually checking the posts. With the ever-increasing volume of data on the platform, manual moderation does not seem a feasible solution in the long run. Hence, platforms are looking toward automatic moderation systems for maintaining civility in their platforms. It has already been observed that Facebook has actively removed a large portion of malicious content from their platforms even before the users report them~\cite{robertson_2020}. However, the limitation is these platforms can detect such abusive content in specific major languages, such as English, Spanish, etc.~\cite{perrigo_2019,Das2021YouTB}. Hence, an effort is required to detect and mitigate offensive/hate speech-language in low resource language. It has been found that Facebook has the highest number of users, and Twitter has the third-highest number of users in India. So it is necessary for these platforms to have moderation systems for Indian languages as well.\par

There is a lot of state-of-the-art hate speech detection research content present in the market, mostly in English languages~\cite{DasNews2020}. To extend the research in other languages, we also study these methods, which detect offensive/ hate content in Hindi, Marathi, Code-Mixed languages using the data in this shared task.

Despite being the third most spoken language, Hindi is always being considered a low resource language because of its mostly typological representation. Marathi is also kind of very low resource language because there is rarely some work present to identify Hate/ Offensive content. Finally, Code-mixed data is also following a current trend because of complexity in writing local languages in universal key inputs. \par \medskip

Earlier, in HASOC 2019\footnote{\url{https://hasocfire.github.io/hasoc/2019/index.html}}  three datasets have been launched to identify Hate and offensive content in English, German and Hindi languages, and in HASOC 2020\footnote{\url{https://hasocfire.github.io/hasoc/2020/index.html}} another dataset has been launched, aiming to identify offensive posts in the code-mixed dataset. Extending the previous work, this time HASOC\cite{hasoc2021mergeoverview,hasoc2021overview}  has introduced two Sub-task, where Sub-task 1 is further divided into two parts. Sub-task 1A focus on Hate speech and Offensive language identification offered for English, Hindi and Marathi. Sub-task 1A is a coarse-grained binary classification in which the posts have to be classified into two classes, namely: Hate and Offensive (HOF) and Non-Hate and offensive (NOT). Sub-task 1B  is a fine-grained classification offered for English and Hindi. Hate-speech and offensive posts from sub-task A are further classified into three categories, Hate speech, offensive and profane. On the other hand, Sub-task 2~\cite{hasoc2021ICHCLoverview} focuses on identifying conversational hate-speech in code-mixed languages. In Sub-task 1B, we participated only in the English language. The definitions of different class labels are given below:

\begin{itemize}
\item {\bf HATE - Hate Speech \cite{mathew2020hatexplain}}: A post is targeting a specific group of people based on their ethnicity, religious beliefs, geographical belonging, race, etc., with malicious intentions of spreading hate or encouraging violence.\bigskip

\item {\bf OFFN - Offensive \cite{6406271} \footnote{\url{https://www.vocabulary.com/dictionary/offensive}}:} Offensive describes rude or hurtful behaviour or a military or sports incursion into an opponent's territory. In any context, "on the offensive" means on the attack.\bigskip

\item {\bf PRFN - Profane \cite{10.1145/3193077.3193078} \footnote{\url{https://www.vocabulary.com/dictionary/profane}}:} A post that expresses deeply offensive behaviour shows a lack of respect, especially for someone's religious beliefs.\bigskip
\end{itemize}

In this paper, we have investigated several Transformer based models for our classification task, which has already been seen to be outperforming the existing baselines and standing as a state-of-the-art model. We perform pre-processing, data sampling, hyper-parameter tuning etc., to build the model. Our models are standing in the \(2nd\) position in Code-Mixed Data set, \(2nd\) position in Hindi two-class classification, \(4th\) in English four-class classification and \(12th\) in English two-class classification. The rest of the paper is organised as follows: Related literature for Hate speech and offensive language detection Section 2. We have discussed the Dataset Description in Section 3; In Section 4, we have presented the System Description. Finally, we have evaluated the experimental setup in Section 5 and the Conclusion in Section 6.

\section{Related Works}

The problem of hate/offensive speech has been studied for a long time in the research community. People were continuously trying to improve the models in order to identify hateful/offensive content more precisely. One of the earliest works that tried to detect hate speech by using lexicon-based features~\cite{6406271Chen}. Although they have provided an efficient framework for future research, their dataset was short for any conclusive evidence. In 2017, Davidson et al. ~\cite{Davidson2017AutomatedHS} contributed a dataset in which thousands of tweets were labelled hate, offensive, and neither. With the classification task of detecting hate/offensive speech present in Tweets in mind. Using this dataset, they then explored how linguistic features such as character and word n-grams affected the performance of a classifier aimed to distinguish the three types of Tweets. Additional features in their classification involved binary and count indicators for hashtags, mentions, retweets, and URLs, as well as
features for the number of characters, words, and syllables in each tweet. The authors found that one of the issues with their best performing models was that they could not distinguish between hate and offensive posts. With the advent of neural networks becoming more accessible and usable for people, many of them tried solutions using these models. 

In 2018, Pitsilis et al.~\cite{Pitsilis2018DetectingOL}, tried deep learning models such as recurrent neural networks (RNNs) to identify the offensive language in English and found that it was quite effective in this task. RNN’s remember the output of each step the model conducts. This approach can capture linguistic context within a text, which is critical to detection. In contrast, RNN’s have been projected to work well with language models, other neural network models, such as CNN. LSTM has had notable success in detecting hate/offensive speech~\cite{Goldberg2015,Sarracn2018HateSD}. 

Although the research on hate/offensive speech detection has been growing rapidly, one of the current issues is that most of the datasets are available in the English language only. Thus, hate/offensive speech in other languages are not detected properly and this could be harmful. This is also a problem for companies like Facebook, which can only detect hate speech in certain languages (English, Spanish, and Mandarin)~\cite{perrigo_2019}. Recently, the research community has begun to focus on hate/offensive language detection in other low resourced languages like Danish~\cite{sigurbergsson-derczynski-2020-offensive}, Greek ~\cite{Pitenis2020OffensiveLI} and Turkish\cite{ltekin2020ACO}.  In the Indian context, the HASOC 2019~\footnote{~\url{https://hasocfire.github.io/hasoc/2019/}} shared Task by Mandal et al.~\cite{Mandl2019OverviewOT} was a significant effort in that direction, where authors created a dataset of hateful and offensive posts in Hindi and English. The best model in this competition used an ensemble of multilingual Transformers, fine-tuned on the given dataset~\cite{Mishra20193IdiotsAH}. In the Dravidian part of HASOC 2020~\footnote{~\url{https://hasocfire.github.io/hasoc/2020/}}, Renjit and Idicula~\cite{Renjit2020} used an ensemble of deep learning and simple neural networks to identify offensive posts in Manglish (Malayalam in the roman font).

Recently, Transformer based~\cite{Vaswani2017AttentionIA} language models such as BERT, m-BERT, XLM-RoBERTa~\cite{Devlin2019BERTPO} are becoming quite popular in several downstream tasks, such as classification, spam detection etc. Previously, it has been already seen these Transformer based models have been outperformed several deep learning models~\cite{mathew2020hatexplain} such as CNN-GRU, LSTM etc. Having observed the superior performance of these Transformer based models, we focus on building these models for our classification problem. 

\section{Dataset Description}

The shared tasks present in this competition are divided into two parts. The datasets have been sampled from Twitter. Subtask-1 offers in English, Hindi with two problems, and Marathi with one problem. The subtasks-2 dataset contains English, Hindi and code-mixed Hindi tweets. Details about the problem statements have been discussed below:

\subsection{Subtask 1A: Identifying Hate, offensive and profane content from the post}

The primary focus of Subtask 1A on Hate speech and Offensive language identification, mainly for English, Hindi and Marathi \cite{gaikwad2021cross}, is coarse-grained binary classification. In Table \ref{tab:Two-class_dataset_statistics} we have presented the dataset statistics on English and Hindi for binary classification.

\subsection{Subtask 1B: Discrimination between Hate, profane and offensive posts}

This subtask is a fine-grained classification of English and Hindi. Mostly Hate-speech and offensive posts from Subtask 1A are further classified as (HATE) Hate speech, (OFFN) Offensive, (PRFN) Profane, (NONE) Non-Hate. In Table \ref{tab:Four-Class-Dataset} we have presented the dataset statistics on English for Four-class classification (We participated for English language only.).

\subsection{Subtask 2: Identification of Conversational Hate-Speech in Code-Mixed Languages (ICHCL)}

A conversational thread can also contain hate and offensive content, which not always can be identified from a single comment or reply to a comment. In this type of situation, context is important to identify the hate or offensive content.\par
\begin{figure*}
  \centering
  \includegraphics[width=0.5\linewidth]{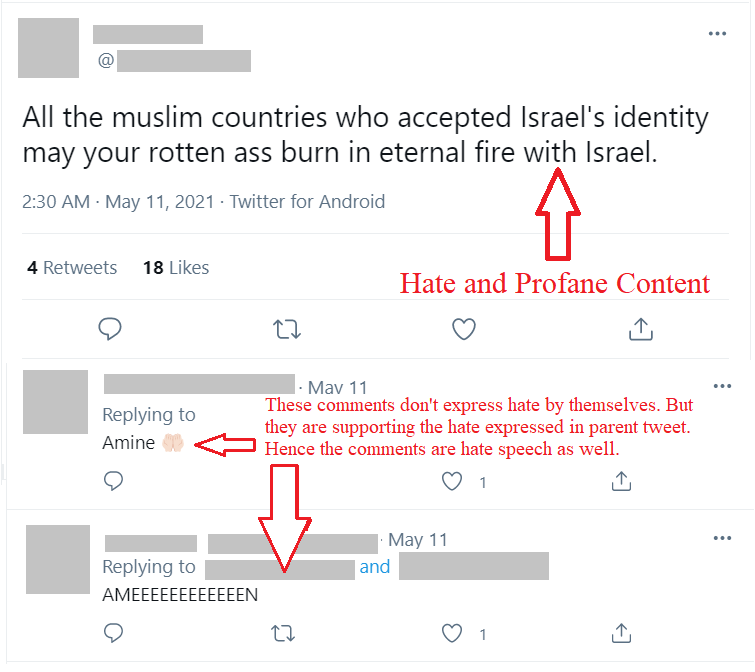}
  \caption{\footnotesize{Example of conversational Hate speech (Image has been taken from HASOC 2021 website)}}
  \label{fig:example}
\end{figure*}

According to Figure \ref{fig:example} the parent tweet is expressing hate and profanity towards Muslim countries regarding the controversy happening in Israel at the time. The two comments on the tweet have written "Amine", which means "truthfully" in Persian,  which supports the hate but with the context of the parent. This sub-task focused on the binary classification of such conversational tweets with tree-structured data into (NOT) Non-Hate-Offensive and (HOF) Hate and Offensive. In Table \ref{tab:Two-class_dataset_statistics} we have presented the dataset statistics of the  code-mixed data for binary classification as well.\par

\begin{table*}
\centering
\begin{tabular}{|c|c|c|c|c|c|c|c|c|} 
\toprule
\multirow{2}{*}{Category} & \multicolumn{2}{c|}{English} & \multicolumn{2}{c|}{Hindi} & \multicolumn{2}{c|}{Marathi} & \multicolumn{2}{c|}{Code-Mixed} \\ 
\cline{2-9}
                            & {Train}& {Test}  & {Train} & {Test} & {Train} & {Test}  & {Train} & {Test}  \\ \midrule
(NOT) Non Hate-Offensive    & {1342} &{483}    & {3161} & {1027}  & {1205} & {418}    & {2899} & {695}      \\ \midrule
(HOF) Hate and Offensive    & {2501} & {798}   & {1433} & {505} & {669} & {207}    & {2841} & {653}       \\ \midrule
\multicolumn{1}{|r|}{Total} & {3843} & {1281}    & {4594} & {1532}  & {1874} & {625}   & {5740} & {1348}      \\ \bottomrule
\end{tabular}
\caption{Two-Class Dataset statistics for languages English, Hindi, Marathi and Code-Mixed}
\label{tab:Two-class_dataset_statistics}
\end{table*}

\begin{table*}
\centering
\begin{tabular}{|c|c|c|}
\toprule
\multirow{2}{*}{Category}                  & \multicolumn{2}{c|}{English} \\ 
\cline{2-3}
                            & {Train} & {Test}   \\ \midrule
(HATE) Hate speech          & {683} & {224}    \\ \midrule
(OFFN) Offensive            & {622} & {195}    \\ \midrule
(PRFN) Profane              & {1196} & {379}   \\ \midrule
(NONE) Non-Hate             & {1342} & {483}   \\ \midrule
\multicolumn{1}{|r|}{Total} & {3843} & {1281}   \\ \bottomrule
\end{tabular}
\caption{Four-Class Dataset statistics for languages English}
\label{tab:Four-Class-Dataset}
\end{table*}

\subsection{Pre-Processing}
While manually going through the data, we found the dataset contains lots of special characters, emoji's, blank spaces, links etc. Mostly custom functions have been used in pre-processing the files, but some libraries were helpful, like "emoji", "nltk" as a baseline. Performed pre-processing steps are:

\begin{itemize}
\item We have replaced all the tagged user names to @user.

\item We have removed all non-alphanumeric characters except full stop and punctuation (| , ?) in Hindi and Marathi. We have kept all the stop words because by that way machine will be able to identify the sequence of characters properly.

\item We have removed emojis, flags and emotions.
\item We have removed all the URLs.
\item We have kept the hashtags because the hashtags contains some contextual information
\end{itemize}

\begin{figure*}
  \centering
  \includegraphics[width=\linewidth]{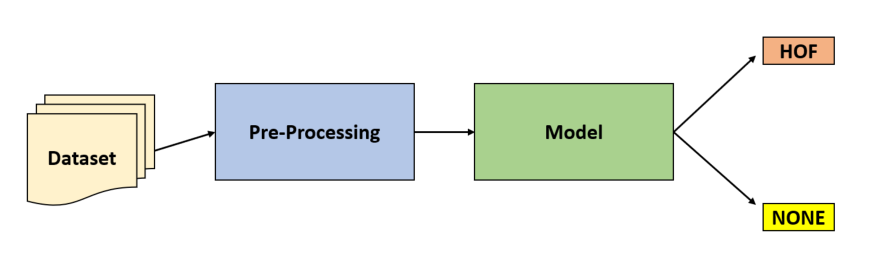}
  \caption{\footnotesize{Design of the proposed methods}}
  \label{fig:violation}
\end{figure*}

\section{System Description}

We have presented our proposed models for Offensive language detection and Hate speech detection in English, Marathi, Hindi and Code-mixed posts. The overall pipeline of the methodology has been represented in Figure ~\ref{fig:violation}. The baseline we have used is Transformer based pre-trained architecture of BERT \cite{Devlin2019BERTPO}. More intuitively, we have used a couple of various versions of BERT, more specifically mBERT \cite{Devlin2019BERTPO} and XLM-Roberta \cite{conneau2020unsupervised}. The beauty of XLM-Roberta is it is trained in an unsupervised manner on the multilingual corpus. XLM-Roberta has achieved state-of-the-art results in most language modelling tasks.

\subsection{Binary Classification}

Most of our task was binary classification problem based on respective embedding. We fine-tuned BERT Transformer and classifier layer on top and used binary target labels for individual classes. We have used this procedure with dehatebert-mono-english \cite{aluru2021deep} and XLM-Roberta 
\cite{conneau2020unsupervised} for English dataset. We have used multilingual BERT (mBERT) and XLM-Roberta for Marathi, Hindi and Code-Mixed classification. Binary cross-entropy loss can be computed for previously mentioned classification task can be mathematically formulated as:

\[Loss = -\sum_{i=1}^{C'=2}t_{i} log (s_{i}) = -t_{1} log(s_{1}) - (1 - t_{1}) log(1 - s_{1})\]

Where it’s assumed that there are two classes: C1 and C2. t1 [0,1] and s1 are the ground truth and the score for C1, and t2=1-t1 and s2=1-s1 are the ground truth and the score for C2. 
\subsection{Multi-class Classification}

In this procedure, we have considered the problem as a Multi-Class classification task. We have fine-tuned the BERT model to get the contextualized embedding by the attention mechanism. We have tried Weighted XLM-Roberta large and weighted dehatebert-mono-english for the Four-Class classification task.

\subsection{Weighted Binary Classification}

The main challenge in any classification problem is the imbalance in data. This imbalance in data may create a bias towards the most present labels, which leads to a decrease in classification performance. According to table~\ref{tab:Two-class_dataset_statistics} it is clearly evident that except code-Mixed data set, there are class imbalances present in the English, Hindi and Marathi dataset. In the English dataset (HOF), Hate and Offensive labels are 46\% more than (NOT) Non-Hate-Offensive class. Similarly, in Hindi (NOT), Non-Hate-Offensive class labels are 54\% more than (HOF) Hate and Offensive class. In Marathi, also (NOT) Non-Hate-Offensive class labels are 44\% more than (HOF) Hate and Offensive class.\par
There is a lot of research has been done in this domain to make the data balance. Oversampling and Undersampling are very much popular data balancing methods, but they have coherent disadvantages also. We tried to implement data balance by using the class weight procedure. Table ~\ref{tab:Class-weight} describes the class weight distribution we have used in order to manage the data imbalance. 

\begin{table}[]
\centering
\begin{tabular}{@{}|c|c|c|l|l|@{}}
\toprule
Task Name & \begin{tabular}[c]{@{}c@{}}(NOT) Non Hate-Offensive\\ Class Weight\end{tabular} & \multicolumn{3}{c|}{\begin{tabular}[c]{@{}c@{}}(HOF) Hate and Offensive\\ Class Weight\end{tabular}} \\ \midrule
English   & 1.4318                                                                          & \multicolumn{3}{c|}{0.7682}                                                                          \\ \midrule
Hindi     & 0.7266                                                                          & \multicolumn{3}{c|}{1.6029}                                                                          \\ \midrule
Marathi   & 0.7775                                                                          & \multicolumn{3}{c|}{1.4005}                                                                          \\ \bottomrule
\end{tabular}
\caption{Normalized Class Weight for Two Class Classification}
\label{tab:Class-weight}
\end{table}

\subsection{Weighted Multi-Class Classification}

In Multi-Class classification also clearly, there is data imbalance present, and we normalized it. It is evident from Table ~\ref{tab:Class-weight} that (HATE) Hate Speech and (OFFN) Offensive counts are quite similar, but they are almost 50\% less than (PRFN) Profane and (NONE) Non-Hate individually. We computed class weight for Multi-Class classification, which is present in Table \ref{tab:Normalized-Multi-Class-Classweight}.

\begin{table}[]
\centering
\begin{tabular}{@{}|c|c|c|c|c|l|@{}}
\toprule
Task Name                & \begin{tabular}[c]{@{}c@{}}(HATE) Hate speech\\ Class Weight\end{tabular} & \begin{tabular}[c]{@{}c@{}}(OFFN) Offensive\\ Class Weight\end{tabular} & \begin{tabular}[c]{@{}c@{}}(PRFN) Profane\\ Class Weight\end{tabular} & \multicolumn{2}{c|}{\begin{tabular}[c]{@{}c@{}}(NONE) Non-Hate\\ Class Weight\end{tabular}} \\ \midrule
\multirow{3}{*}{English} & \multirow{3}{*}{1.4066}                                                   & \multirow{3}{*}{1.5446}                                                 & \multirow{3}{*}{0.8033}                                               & \multicolumn{2}{c|}{\multirow{3}{*}{0.7159}}                                                \\
                         &                                                                           &                                                                         &                                                                       & \multicolumn{2}{c|}{}                                                                       \\
                         &                                                                           &                                                                         &                                                                       & \multicolumn{2}{c|}{}                                                                       \\ \bottomrule
\end{tabular}
\caption{Normalized Multi-Class Classweight}
\label{tab:Normalized-Multi-Class-Classweight}
\end{table}

\subsection{Tuning Parameters}

For all the models presented here, we have pre-train on the target dataset for 20 epochs in order to capture the semantics. Along with that, we fine-tuned weighted and unweighted using cross-entropy loss functions\cite{10.1145/1102351.1102422}. We have used HuggingFace\cite{wolf2020huggingfaces} and PyTorch \cite{paszke2019pytorch}. Initial phases, we have used Adam optimizer\cite{loshchilov2019decoupled} with an initial learning rate as 2e-5. We have not used early stopping while training.

\section{Results}

Our observation was among most of the individual Transformer based BERT models, and the best performance was coming using XLM-Roberta-large (XLMR-large) in English Two-Class and Four-Class dataset and Marathi dataset. In contrast, we are getting the best performance in Code-Mixed dataset by using Custom XLM-Roberta-large. In the case of the Hindi dataset, mBERT is giving the best performance. The beauty of XLM-Roberta is that it has been pre-trained on the parallel corpus. We have noted that the performance of XLM-Roberta-large is very much consistent with most of the regional languages.\par
While achieving the performance scores. We have used multiple random seeds and have observed that performance was heavily getting impacted for different seeds. It has been observed that while using mBERT, the performance varied 6-7\% across our experimented languages. In the case of XLM-Roberta models, the performance was mostly the same and, it varied a maximum 1-2\%. Table \ref{tab:Result-Two-Class} shows the performance of XLMR-base, XLMR-large and mBERT-base for Two Class classification results. We have shown the classification results for Four-Class classification in Table \ref{tab:Result-Four-Class}. \par
This team have not actively participated in the competition for the Marathi dataset, but later post-competition implemented all the transformer based models mentioned in this paper and found Macro F1 as 0.8756, which matches with the 3rd rank holder team from the competition.
\begin{table*}
\centering
\begin{tabular}{|c|c|c|c|c|} 
\toprule
\multirow{2}{*} {Classifiers} & English         & Hindi                  & Code-Mixed      & Marathi\\ 
\cline{2-5}
            & Macro F1        & Macro F1               & Macro F1        & Macro F1\\ \midrule
XLMR-base   & 0.7834          & 0.6862                   & 0.6456          & 0.8133\\ \midrule
XLMR-large  & \textbf{0.8006} & 0.7112           & \textbf{0.7107}  & \textbf{0.8756}\\ \midrule
mBERT-base  & 0.7328          & \textbf{0.7797}          & 0.6277          & 0.8611\\ \midrule
Indic-BERT  & 0.7002          & 0.6323                  & 0.5912          & 0.8176\\ \midrule
dehate-BERT & 0.7811          & 0.6533                    & 0.6377    & 0.7550\\ \midrule
\textbf{Submission Name} & "Bestn"     & "T2"      & "Context 1"   &  - \\  \bottomrule
\end{tabular}
\caption{Two-Class Classification Result}
\label{tab:Result-Two-Class}
\end{table*}

\begin{table*}
\centering
\begin{tabular}{|c|c|c|c|c|}
\toprule
\multirow{2}{*}{Classifiers} & \multicolumn{4}{c|}{English}         \\ 
\cline{2-5}
           & \multicolumn{4}{c|}{Macro F1}        \\ \midrule
XLMR-base  & \multicolumn{4}{c|}{0.5824}          \\ \midrule
XLMR-large & \multicolumn{4}{c|}{\textbf{0.6447}} \\ \midrule
mBERT-base & \multicolumn{4}{c|}{0.5443}          \\ \midrule
Indic-BERT  & \multicolumn{4}{c|}{0.5119} \\ \midrule
dehate-BERT & \multicolumn{4}{c|}{0.4845} \\ \midrule
\textbf{Submission Name} & \multicolumn{4}{c|}{"Final"} \\ \bottomrule
\end{tabular}
\caption{Four-Class Classification Result}
\label{tab:Result-Four-Class}
\end{table*}

\section{Conclusion}

In this shared task, we have compared and evaluated multiple Transformer-based architectures and discovered that XLM-Roberta-large mainly performs better than other Transformer-based models. However, performance varies based on a random seed. It has been observed that by changing random seed, XLM-Roberta performance was impacted less than other Transformer-based models. So, some of the actions will be to identify this observation and speculate the reason behind it. We have also used a couple of Transformer based models like IndicBERT and dehateBERT but was not getting enough raising performance compared with XLMRobeta and mBERT. Our immediate next step will be to investigate the reasons behind the lower performance of IndicBERT and dehateBERT, as IndicBERT is specifically pretrained with Indian languages and dehateBERT is an already fine-tuned model on the hate speech dataset.



\bibliography{sample-ceur}

\begin{thebibliography}{32}
\expandafter\ifx\csname natexlab\endcsname\relax\def\natexlab#1{#1}\fi
\providecommand{\url}[1]{\texttt{#1}}
\providecommand{\href}[2]{#2}
\providecommand{\path}[1]{#1}
\providecommand{\DOIprefix}{doi:}
\providecommand{\ArXivprefix}{arXiv:}
\providecommand{\URLprefix}{URL: }
\providecommand{\Pubmedprefix}{pmid:}
\providecommand{\doi}[1]{\href{http://dx.doi.org/#1}{\path{#1}}}
\providecommand{\Pubmed}[1]{\href{pmid:#1}{\path{#1}}}
\providecommand{\bibinfo}[2]{#2}
\ifx\xfnm\relax \def\xfnm[#1]{\unskip,\space#1}\fi
\bibitem[{Mathew et~al.(2019)Mathew, Dutt, Goyal, and
  Mukherjee}]{hate-speech-websci-19}
\bibinfo{author}{B.~Mathew}, \bibinfo{author}{R.~Dutt},
  \bibinfo{author}{P.~Goyal}, \bibinfo{author}{A.~Mukherjee},
\newblock \bibinfo{title}{Spread of hate speech in online social media},
\newblock in: \bibinfo{booktitle}{Proceedings of WebSci},
  \bibinfo{publisher}{ACM}, \bibinfo{year}{2019}.
\bibitem[{Newton(2019)}]{newton_2019}
\bibinfo{author}{C.~Newton}, \bibinfo{title}{The terror queue},
  \bibinfo{year}{2019}. \URLprefix
  \url{https://www.theverge.com/2019/12/16/21021005/google-youtube-moderators-ptsd-accenture-violent-disturbing-content-interviews-video}.
\bibitem[{Robertson(2020)}]{robertson_2020}
\bibinfo{author}{A.~Robertson}, \bibinfo{title}{Facebook says ai has fueled a
  hate speech crackdown}, \bibinfo{year}{2020}. \URLprefix
  \url{https://www.theverge.com/2020/11/19/21575139/facebook-moderation-ai-hate-speech}.
\bibitem[{Perrigo(2019)}]{perrigo_2019}
\bibinfo{author}{B.~Perrigo}, \bibinfo{title}{Facebook's hate speech algorithms
  leave out some languages}, \bibinfo{year}{2019}. \URLprefix
  \url{https://time.com/5739688/facebook-hate-speech-languages/}.
\bibitem[{Das et~al.(2021)Das, Saha, Dutt, Goyal, Mukherjee, and
  Mathew}]{Das2021YouTB}
\bibinfo{author}{M.~Das}, \bibinfo{author}{P.~Saha}, \bibinfo{author}{R.~Dutt},
  \bibinfo{author}{P.~Goyal}, \bibinfo{author}{A.~Mukherjee},
  \bibinfo{author}{B.~Mathew},
\newblock \bibinfo{title}{You too brutus! trapping hateful users in social
  media: Challenges, solutions \& insights},
\newblock \bibinfo{year}{2021}, pp. \bibinfo{pages}{79--89}.
  \DOIprefix\doi{10.1145/3465336.3475106}.
\bibitem[{Das et~al.(2020)Das, Mathew, Saha, Goyal, and
  Mukherjee}]{DasNews2020}
\bibinfo{author}{M.~Das}, \bibinfo{author}{B.~Mathew},
  \bibinfo{author}{P.~Saha}, \bibinfo{author}{P.~Goyal},
  \bibinfo{author}{A.~Mukherjee},
\newblock \bibinfo{title}{Hate speech in online social media},
\newblock \bibinfo{journal}{ACM SIGWEB Newsletter}  (\bibinfo{year}{2020})
  \bibinfo{pages}{1--8}. \DOIprefix\doi{10.1145/3427478.3427482}.
\bibitem[{Modha et~al.(2021)Modha, Mandl, Shahi, Madhu, Satapara, Ranasinghe,
  and Zampieri}]{hasoc2021mergeoverview}
\bibinfo{author}{S.~Modha}, \bibinfo{author}{T.~Mandl}, \bibinfo{author}{G.~K.
  Shahi}, \bibinfo{author}{H.~Madhu}, \bibinfo{author}{S.~Satapara},
  \bibinfo{author}{T.~Ranasinghe}, \bibinfo{author}{M.~Zampieri},
\newblock \bibinfo{title}{Overview of the hasoc subtrack at fire 2021: Hate
  speech and offensive content identification in english and indo-aryan
  languages and conversational hate speech},
\newblock in: \bibinfo{booktitle}{{FIRE} 2021: Forum for Information Retrieval
  Evaluation, Virtual Event, 13th-17th December 2021},
  \bibinfo{publisher}{ACM}, \bibinfo{year}{2021}.
\bibitem[{Mandl et~al.(2021)Mandl, Modha, Shahi, Madhu, Satapara, Majumder,
  Schäfer, Ranasinghe, Zampieri, Nandini, and Jaiswal}]{hasoc2021overview}
\bibinfo{author}{T.~Mandl}, \bibinfo{author}{S.~Modha}, \bibinfo{author}{G.~K.
  Shahi}, \bibinfo{author}{H.~Madhu}, \bibinfo{author}{S.~Satapara},
  \bibinfo{author}{P.~Majumder}, \bibinfo{author}{J.~Schäfer},
  \bibinfo{author}{T.~Ranasinghe}, \bibinfo{author}{M.~Zampieri},
  \bibinfo{author}{D.~Nandini}, \bibinfo{author}{A.~K. Jaiswal},
\newblock \bibinfo{title}{{Overview of the HASOC subtrack at FIRE 2021: Hate
  Speech and Offensive Content Identification in English and Indo-Aryan
  Languages}},
\newblock in: \bibinfo{booktitle}{Working Notes of FIRE 2021 - Forum for
  Information Retrieval Evaluation}, \bibinfo{publisher}{CEUR},
  \bibinfo{year}{2021}. \URLprefix \url{http://ceur-ws.org/}.
\bibitem[{Satapara et~al.(2021)Satapara, Modha, Mandl, Madhu, and
  Majumder}]{hasoc2021ICHCLoverview}
\bibinfo{author}{S.~Satapara}, \bibinfo{author}{S.~Modha},
  \bibinfo{author}{T.~Mandl}, \bibinfo{author}{H.~Madhu},
  \bibinfo{author}{P.~Majumder},
\newblock \bibinfo{title}{{ Overview of the HASOC Subtrack at FIRE 2021:
  Conversational Hate Speech Detection in Code-mixed language }},
\newblock in: \bibinfo{booktitle}{Working Notes of FIRE 2021 - Forum for
  Information Retrieval Evaluation}, \bibinfo{publisher}{CEUR},
  \bibinfo{year}{2021}.
\bibitem[{Mathew et~al.(2020)Mathew, Saha, Yimam, Biemann, Goyal, and
  Mukherjee}]{mathew2020hatexplain}
\bibinfo{author}{B.~Mathew}, \bibinfo{author}{P.~Saha}, \bibinfo{author}{S.~M.
  Yimam}, \bibinfo{author}{C.~Biemann}, \bibinfo{author}{P.~Goyal},
  \bibinfo{author}{A.~Mukherjee}, \bibinfo{title}{Hatexplain: A benchmark
  dataset for explainable hate speech detection}, \bibinfo{year}{2020}.
  \href{http://arxiv.org/abs/2012.10289}{{\tt arXiv:2012.10289}}.
\bibitem[{Chen et~al.(2012)Chen, Zhou, Zhu, and Xu}]{6406271}
\bibinfo{author}{Y.~Chen}, \bibinfo{author}{Y.~Zhou}, \bibinfo{author}{S.~Zhu},
  \bibinfo{author}{H.~Xu},
\newblock \bibinfo{title}{Detecting offensive language in social media to
  protect adolescent online safety},
\newblock in: \bibinfo{booktitle}{2012 International Conference on Privacy,
  Security, Risk and Trust and 2012 International Confernece on Social
  Computing}, \bibinfo{year}{2012}, pp. \bibinfo{pages}{71--80}.
  \DOIprefix\doi{10.1109/SocialCom-PASSAT.2012.55}.
\bibitem[{Teh et~al.(2018)Teh, Cheng, and Chee}]{10.1145/3193077.3193078}
\bibinfo{author}{P.~L. Teh}, \bibinfo{author}{C.-B. Cheng},
  \bibinfo{author}{W.~M. Chee},
\newblock \bibinfo{title}{Identifying and categorising profane words in hate
  speech},
\newblock in: \bibinfo{booktitle}{Proceedings of the 2nd International
  Conference on Compute and Data Analysis}, ICCDA 2018,
  \bibinfo{publisher}{Association for Computing Machinery},
  \bibinfo{address}{New York, NY, USA}, \bibinfo{year}{2018}, p.
  \bibinfo{pages}{65–69}. \URLprefix
  \url{https://doi.org/10.1145/3193077.3193078}.
  \DOIprefix\doi{10.1145/3193077.3193078}.
\bibitem[{Chen et~al.(2012)Chen, Zhou, Zhu, and Xu}]{6406271Chen}
\bibinfo{author}{Y.~Chen}, \bibinfo{author}{Y.~Zhou}, \bibinfo{author}{S.~Zhu},
  \bibinfo{author}{H.~Xu},
\newblock \bibinfo{title}{Detecting offensive language in social media to
  protect adolescent online safety},
\newblock in: \bibinfo{booktitle}{2012 International Conference on Privacy,
  Security, Risk and Trust and 2012 International Confernece on Social
  Computing}, \bibinfo{year}{2012}, pp. \bibinfo{pages}{71--80}.
  \DOIprefix\doi{10.1109/SocialCom-PASSAT.2012.55}.
\bibitem[{Davidson et~al.(2017)Davidson, Warmsley, Macy, and
  Weber}]{Davidson2017AutomatedHS}
\bibinfo{author}{T.~Davidson}, \bibinfo{author}{D.~Warmsley},
  \bibinfo{author}{M.~Macy}, \bibinfo{author}{I.~Weber},
\newblock \bibinfo{title}{Automated hate speech detection and the problem of
  offensive language},
\newblock in: \bibinfo{booktitle}{ICWSM}, \bibinfo{year}{2017}.
\bibitem[{Pitsilis et~al.(2018)Pitsilis, Ramampiaro, and
  Langseth}]{Pitsilis2018DetectingOL}
\bibinfo{author}{G.~K. Pitsilis}, \bibinfo{author}{H.~Ramampiaro},
  \bibinfo{author}{H.~Langseth},
\newblock \bibinfo{title}{Detecting offensive language in tweets using deep
  learning},
\newblock \bibinfo{journal}{ArXiv} \bibinfo{volume}{abs/1801.04433}
  (\bibinfo{year}{2018}).
\bibitem[{Goldberg(2015)}]{Goldberg2015}
\bibinfo{author}{Y.~Goldberg},
\newblock \bibinfo{title}{A primer on neural network models for natural
  language processing},
\newblock \bibinfo{journal}{Journal of Artificial Intelligence Research}
  \bibinfo{volume}{57} (\bibinfo{year}{2015}).
  \DOIprefix\doi{10.1613/jair.4992}.
\bibitem[{la~Pe{\~n}a~Sarrac{\'e}n et~al.(2018)la~Pe{\~n}a~Sarrac{\'e}n, Pons,
  Mu{\~n}iz-Cuza, and Rosso}]{Sarracn2018HateSD}
\bibinfo{author}{G.~L.~D. la~Pe{\~n}a~Sarrac{\'e}n}, \bibinfo{author}{R.~G.
  Pons}, \bibinfo{author}{C.~E. Mu{\~n}iz-Cuza}, \bibinfo{author}{P.~Rosso},
\newblock \bibinfo{title}{Hate speech detection using attention-based lstm},
\newblock in: \bibinfo{booktitle}{EVALITA@CLiC-it}, \bibinfo{year}{2018}.
\bibitem[{Sigurbergsson and
  Derczynski(2020)}]{sigurbergsson-derczynski-2020-offensive}
\bibinfo{author}{G.~I. Sigurbergsson}, \bibinfo{author}{L.~Derczynski},
\newblock \bibinfo{title}{Offensive language and hate speech detection for
  {D}anish},
\newblock in: \bibinfo{booktitle}{Proceedings of the 12th Language Resources
  and Evaluation Conference}, \bibinfo{publisher}{European Language Resources
  Association}, \bibinfo{address}{Marseille, France}, \bibinfo{year}{2020}, pp.
  \bibinfo{pages}{3498--3508}. \URLprefix
  \url{https://aclanthology.org/2020.lrec-1.430}.
\bibitem[{Pitenis et~al.(2020)Pitenis, Zampieri, and
  Ranasinghe}]{Pitenis2020OffensiveLI}
\bibinfo{author}{Z.~Pitenis}, \bibinfo{author}{M.~Zampieri},
  \bibinfo{author}{T.~Ranasinghe},
\newblock \bibinfo{title}{Offensive language identification in greek},
\newblock in: \bibinfo{booktitle}{LREC}, \bibinfo{year}{2020}.
\bibitem[{Çagri Ç{\"o}ltekin(2020)}]{ltekin2020ACO}
\bibinfo{author}{Çagri Ç{\"o}ltekin},
\newblock \bibinfo{title}{A corpus of turkish offensive language on social
  media},
\newblock in: \bibinfo{booktitle}{LREC}, \bibinfo{year}{2020}.
\bibitem[{Mandl et~al.(2019)Mandl, Modha, Majumder, Patel, Dave, Mandalia, and
  Patel}]{Mandl2019OverviewOT}
\bibinfo{author}{T.~Mandl}, \bibinfo{author}{S.~Modha},
  \bibinfo{author}{P.~Majumder}, \bibinfo{author}{D.~Patel},
  \bibinfo{author}{M.~Dave}, \bibinfo{author}{C.~Mandalia},
  \bibinfo{author}{A.~Patel},
\newblock \bibinfo{title}{Overview of the hasoc track at fire 2019: Hate speech
  and offensive content identification in indo-european languages},
\newblock \bibinfo{journal}{Proceedings of the 11th Forum for Information
  Retrieval Evaluation}  (\bibinfo{year}{2019}).
\bibitem[{Mishra(2019)}]{Mishra20193IdiotsAH}
\bibinfo{author}{S.~Mishra},
\newblock \bibinfo{title}{3idiots at hasoc 2019: Fine-tuning transformer neural
  networks for hate speech identification in indo-european languages},
\newblock in: \bibinfo{booktitle}{FIRE}, \bibinfo{year}{2019}.
\bibitem[{Renjit and Idicula(2020)}]{Renjit2020}
\bibinfo{author}{S.~Renjit}, \bibinfo{author}{S.~M. Idicula},
\newblock \bibinfo{title}{Cusatnlp@hasoc-dravidian-codemix-fire2020:identifying
  offensive language from manglishtweets},
\newblock \bibinfo{journal}{ArXiv} \bibinfo{volume}{abs/2010.08756}
  (\bibinfo{year}{2020}).
\bibitem[{Vaswani et~al.(2017)Vaswani, Shazeer, Parmar, Uszkoreit, Jones,
  Gomez, Kaiser, and Polosukhin}]{Vaswani2017AttentionIA}
\bibinfo{author}{A.~Vaswani}, \bibinfo{author}{N.~M. Shazeer},
  \bibinfo{author}{N.~Parmar}, \bibinfo{author}{J.~Uszkoreit},
  \bibinfo{author}{L.~Jones}, \bibinfo{author}{A.~N. Gomez},
  \bibinfo{author}{L.~Kaiser}, \bibinfo{author}{I.~Polosukhin},
\newblock \bibinfo{title}{Attention is all you need},
\newblock \bibinfo{journal}{ArXiv} \bibinfo{volume}{abs/1706.03762}
  (\bibinfo{year}{2017}).
\bibitem[{Devlin et~al.(2019)Devlin, Chang, Lee, and
  Toutanova}]{Devlin2019BERTPO}
\bibinfo{author}{J.~Devlin}, \bibinfo{author}{M.-W. Chang},
  \bibinfo{author}{K.~Lee}, \bibinfo{author}{K.~Toutanova},
\newblock \bibinfo{title}{Bert: Pre-training of deep bidirectional transformers
  for language understanding},
\newblock in: \bibinfo{booktitle}{NAACL}, \bibinfo{year}{2019}.
\bibitem[{Gaikwad et~al.(2021)Gaikwad, Ranasinghe, Zampieri, and
  Homan}]{gaikwad2021cross}
\bibinfo{author}{S.~Gaikwad}, \bibinfo{author}{T.~Ranasinghe},
  \bibinfo{author}{M.~Zampieri}, \bibinfo{author}{C.~M. Homan},
\newblock \bibinfo{title}{Cross-lingual offensive language identification for
  low resource languages: The case of marathi},
\newblock in: \bibinfo{booktitle}{Proceedings of RANLP}, \bibinfo{year}{2021}.
\bibitem[{Conneau et~al.(2020)Conneau, Khandelwal, Goyal, Chaudhary, Wenzek,
  Guzmán, Grave, Ott, Zettlemoyer, and Stoyanov}]{conneau2020unsupervised}
\bibinfo{author}{A.~Conneau}, \bibinfo{author}{K.~Khandelwal},
  \bibinfo{author}{N.~Goyal}, \bibinfo{author}{V.~Chaudhary},
  \bibinfo{author}{G.~Wenzek}, \bibinfo{author}{F.~Guzmán},
  \bibinfo{author}{E.~Grave}, \bibinfo{author}{M.~Ott},
  \bibinfo{author}{L.~Zettlemoyer}, \bibinfo{author}{V.~Stoyanov},
  \bibinfo{title}{Unsupervised cross-lingual representation learning at scale},
  \bibinfo{year}{2020}. \href{http://arxiv.org/abs/1911.02116}{{\tt
  arXiv:1911.02116}}.
\bibitem[{Aluru et~al.(2021)Aluru, Mathew, Saha, and Mukherjee}]{aluru2021deep}
\bibinfo{author}{S.~S. Aluru}, \bibinfo{author}{B.~Mathew},
  \bibinfo{author}{P.~Saha}, \bibinfo{author}{A.~Mukherjee},
\newblock \bibinfo{title}{A deep dive into multilingual hate speech
  classification},
\newblock in: \bibinfo{booktitle}{Machine Learning and Knowledge Discovery in
  Databases. Applied Data Science and Demo Track: European Conference, ECML
  PKDD 2020, Ghent, Belgium, September 14--18, 2020, Proceedings, Part V},
  \bibinfo{organization}{Springer International Publishing},
  \bibinfo{year}{2021}, pp. \bibinfo{pages}{423--439}.
\bibitem[{Mannor et~al.(2005)Mannor, Peleg, and
  Rubinstein}]{10.1145/1102351.1102422}
\bibinfo{author}{S.~Mannor}, \bibinfo{author}{D.~Peleg},
  \bibinfo{author}{R.~Rubinstein},
\newblock \bibinfo{title}{The cross entropy method for classification},
\newblock in: \bibinfo{booktitle}{Proceedings of the 22nd International
  Conference on Machine Learning}, ICML '05, \bibinfo{publisher}{Association
  for Computing Machinery}, \bibinfo{address}{New York, NY, USA},
  \bibinfo{year}{2005}, p. \bibinfo{pages}{561–568}. \URLprefix
  \url{https://doi.org/10.1145/1102351.1102422}.
  \DOIprefix\doi{10.1145/1102351.1102422}.
\bibitem[{Wolf et~al.(2020)Wolf, Debut, Sanh, Chaumond, Delangue, Moi, Cistac,
  Rault, Louf, Funtowicz, Davison, Shleifer, von Platen, Ma, Jernite, Plu, Xu,
  Scao, Gugger, Drame, Lhoest, and Rush}]{wolf2020huggingfaces}
\bibinfo{author}{T.~Wolf}, \bibinfo{author}{L.~Debut},
  \bibinfo{author}{V.~Sanh}, \bibinfo{author}{J.~Chaumond},
  \bibinfo{author}{C.~Delangue}, \bibinfo{author}{A.~Moi},
  \bibinfo{author}{P.~Cistac}, \bibinfo{author}{T.~Rault},
  \bibinfo{author}{R.~Louf}, \bibinfo{author}{M.~Funtowicz},
  \bibinfo{author}{J.~Davison}, \bibinfo{author}{S.~Shleifer},
  \bibinfo{author}{P.~von Platen}, \bibinfo{author}{C.~Ma},
  \bibinfo{author}{Y.~Jernite}, \bibinfo{author}{J.~Plu},
  \bibinfo{author}{C.~Xu}, \bibinfo{author}{T.~L. Scao},
  \bibinfo{author}{S.~Gugger}, \bibinfo{author}{M.~Drame},
  \bibinfo{author}{Q.~Lhoest}, \bibinfo{author}{A.~M. Rush},
  \bibinfo{title}{Huggingface's transformers: State-of-the-art natural language
  processing}, \bibinfo{year}{2020}.
  \href{http://arxiv.org/abs/1910.03771}{{\tt arXiv:1910.03771}}.
\bibitem[{Paszke et~al.(2019)Paszke, Gross, Massa, Lerer, Bradbury, Chanan,
  Killeen, Lin, Gimelshein, Antiga, Desmaison, Köpf, Yang, DeVito, Raison,
  Tejani, Chilamkurthy, Steiner, Fang, Bai, and Chintala}]{paszke2019pytorch}
\bibinfo{author}{A.~Paszke}, \bibinfo{author}{S.~Gross},
  \bibinfo{author}{F.~Massa}, \bibinfo{author}{A.~Lerer},
  \bibinfo{author}{J.~Bradbury}, \bibinfo{author}{G.~Chanan},
  \bibinfo{author}{T.~Killeen}, \bibinfo{author}{Z.~Lin},
  \bibinfo{author}{N.~Gimelshein}, \bibinfo{author}{L.~Antiga},
  \bibinfo{author}{A.~Desmaison}, \bibinfo{author}{A.~Köpf},
  \bibinfo{author}{E.~Yang}, \bibinfo{author}{Z.~DeVito},
  \bibinfo{author}{M.~Raison}, \bibinfo{author}{A.~Tejani},
  \bibinfo{author}{S.~Chilamkurthy}, \bibinfo{author}{B.~Steiner},
  \bibinfo{author}{L.~Fang}, \bibinfo{author}{J.~Bai},
  \bibinfo{author}{S.~Chintala}, \bibinfo{title}{Pytorch: An imperative style,
  high-performance deep learning library}, \bibinfo{year}{2019}.
  \href{http://arxiv.org/abs/1912.01703}{{\tt arXiv:1912.01703}}.
\bibitem[{Loshchilov and Hutter(2019)}]{loshchilov2019decoupled}
\bibinfo{author}{I.~Loshchilov}, \bibinfo{author}{F.~Hutter},
  \bibinfo{title}{Decoupled weight decay regularization}, \bibinfo{year}{2019}.
  \href{http://arxiv.org/abs/1711.05101}{{\tt arXiv:1711.05101}}.

\end{thebibliography}

\appendix

\end{document}